\documentclass[conference]{IEEEtran}
\IEEEoverridecommandlockouts
\usepackage{cite}
\usepackage{amsmath,amssymb,amsfonts}
\usepackage{algorithm}
\usepackage{algorithmic}
\usepackage{graphicx}
\usepackage{textcomp}
\usepackage{xcolor}

\usepackage{tabularx}
\usepackage{subcaption}
\usepackage{multirow}
\usepackage{makecell}
\usepackage{float}

\usepackage{tikz}
\usepackage{textcomp}
\usepackage[hidelinks]{hyperref}

\newcommand\copyrighttext{%
  \footnotesize \textcopyright 2012 IEEE. Personal use of this material is permitted.
  Permission from IEEE must be obtained for all other uses, in any current or future
  media, including reprinting/republishing this material for advertising or promotional
  purposes, creating new collective works, for resale or redistribution to servers or
  lists, or reuse of any copyrighted component of this work in other works.
  DOI: \href{https://doi.org/10.1109/BigComp51126.2021.00039}{10.1109/BigComp51126.2021.00039}}
\newcommand\copyrightnotice{%
\begin{tikzpicture}[remember picture,overlay]
\node[anchor=south,yshift=10] at (current page.south) {\fbox{\parbox{\dimexpr\textwidth-\fboxsep-\fboxrule\relax}{\copyrighttext}}};
\end{tikzpicture}%
}

\def\BibTeX{{\rm B\kern-.05em{\sc i\kern-.025em b}\kern-.08em
    T\kern-.1667em\lower.7ex\hbox{E}\kern-.125emX}}

\graphicspath{{./image/}} % Figure path

\begin{document}

\title{Federated Learning based Energy Demand Prediction with Clustered Aggregation \\
\thanks{This work was partially supported by Institute of Information \& communications Technology Planning \& Evaluation (IITP) grant funded by the Korea government(MSIT) (No.2019-0-01287, Evolvable Deep Learning Model Generation Platform for Edge Computing) and by the National Research Foundation of Korea(NRF) grant funded by the Korea government(MSIT) (No. No. 2020R1A4A1018607). *Dr. CS Hong is the corresponding author.}
}

%\author{
%\IEEEauthorblockN{1\textsuperscript{st} Ye Lin Tun}
%\IEEEauthorblockA{\textit{dept. name of organization (of Aff.)}\\
%\and
%\IEEEauthorblockN{2\textsuperscript{nd} Kyi Thar}
%\and
%\IEEEauthorblockN{3\textsuperscript{rd} Chu Myaet Thwal}
%\and
%\IEEEauthorblockN{4\textsuperscript{th} Choong Seon Hong}
%\and
%\and

\author{
	\IEEEauthorblockN{Ye Lin Tun, Kyi Thar, Chu Myaet Thwal, Choong Seon Hong}
	\IEEEauthorblockA{\textit{Department of Computer Science and Engineering}\\
		\textit{Kyung Hee University}\\
		Yongin-si, 17104, Republic of Korea \\
		\{yelintun, kyithar, chumyaet, cshong\}@khu.ac.kr}
}

\maketitle
\copyrightnotice

\begin{abstract}

To reduce negative environmental impacts, power stations and energy grids need to optimize the resources required for power production. Thus, predicting the energy consumption of clients is becoming an important part of every energy management system. Energy usage information collected by the clients' smart homes can be used to train a deep neural network to predict the future energy demand. Collecting data from a large number of distributed clients for centralized model training is expensive in terms of communication resources. To take advantage of distributed data in edge systems, centralized training can be replaced by federated learning where each client only needs to upload model updates produced by training on its local data. These model updates are aggregated into a single global model by the server. But since different clients can have different attributes, model updates can have diverse weights and as a result, it can take a long time for the aggregated global model to converge. To speed up the convergence process, we can apply clustering to group clients based on their properties and aggregate model updates from the same cluster together to produce a cluster specific global model. In this paper, we propose a recurrent neural network based energy demand predictor, trained with federated learning on clustered clients to take advantage of distributed data and speed up the convergence process.

\end{abstract}


\section{Introduction}
\begin{figure*}[htbp]
	\centering
	\includegraphics[height=4.5cm, keepaspectratio]{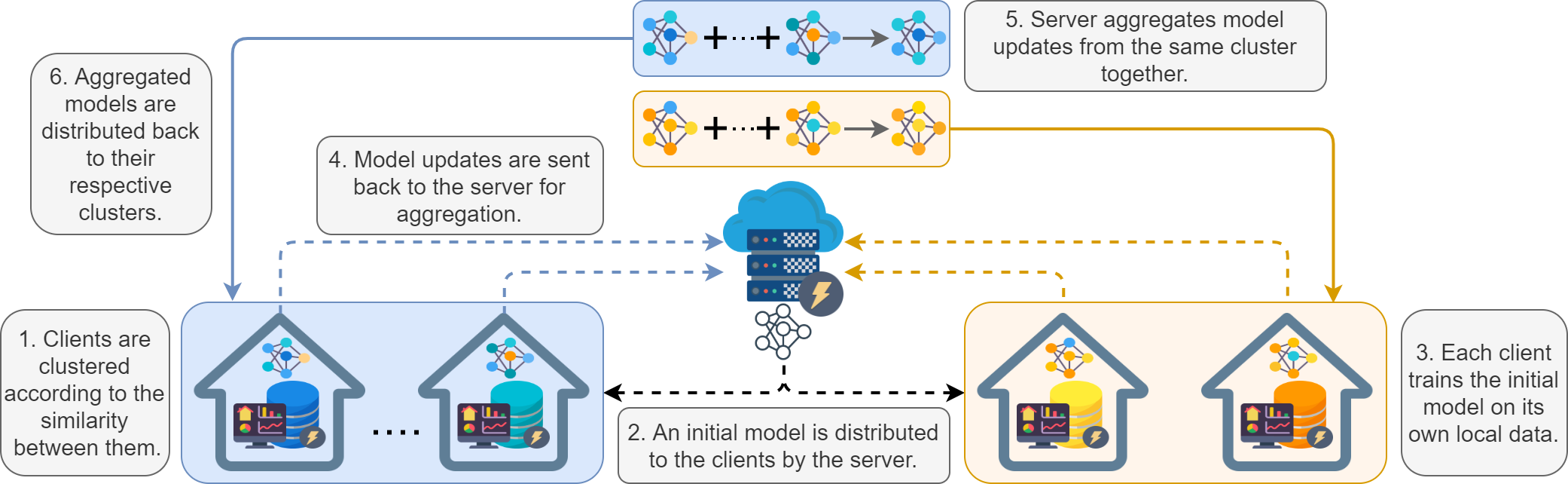}
	\caption{Federated Learning and Clustering based Model Training Process}
	\label{fig:system_model}
\end{figure*}
Electrical energy is an essential component for a single household to all industrial and manufacturing systems \cite{ISMETUGURSAL2014783}. To effectively utilize the available natural resources, power plants need to efficiently produce and distribute energy based on clients’ demand. Nowadays, smart home systems can monitor and log energy usage statistics from various sources across the house. This highly granular information can be valuable for energy companies to effectively manage the power production and distribution operations.

During recent years, increasing computing capabilities and data explosion has led to a significant performance increase for data-driven predictor models \cite{AMASYALI20181192}. Many existing studies on energy consumption modeling \cite{Berriel2017MonthlyEC,Kim2019,DONG2005545,JETCHEVA2014214,EDWARDS2012591} utilize data-driven methods, including  deep neural networks (DNN), decision trees, support vector machines (SVM), and other machine learning methods. In these previous methods, data is centralized for training the data-driven model.

Predicting the future energy demand of a client based on a series of past energy usage data is a time series regression task. Recurrent neural networks (RNNs) are a type of neural network that can accept a series of values as the input to predict the output \cite{Hewamalage2019RecurrentNN} and they are suitable for tackling time series predictive problems. In our system, the RNN model is chosen for predicting the future energy demand of a client given past energy usage data.
  
In a traditional neural network training pipeline, clients’ data are collected onto a central server where the model is trained. Transmitting data from the clients onto a server is expensive in terms of communication cost, and adapting the model to new usage patterns in the future would require constant re-transmission of newly obtained data. However, most of the existing works \cite{Berriel2017MonthlyEC,Kim2019,DONG2005545,JETCHEVA2014214,EDWARDS2012591} only consider the centralized training approach where data transmission is essential.

To overcome these challenges, we can apply federated learning \cite{Konecn2016FederatedLS,Konecn2016FederatedOD} where clients can collaboratively train a deep neural network on their own local data without needing to centralize. However, training a neural network with federated learning usually suffers from the non-IID data problem \cite{Li2020FederatedLC}, where clients contain data distributions that are diverse from each other. As a result, the global model created by aggregating different client model updates has a poor convergence rate and performance. Clustering clients with similar properties is a promising way to alleviated this problem. Model updates produced by the clients of the same cluster can be aggregated together to create a specific global model for that cluster. In this work, we integrated the clustered aggregation in the federated training process of our RNN model, by grouping different model updates based on available client attributes. 
%
%
%\begin{figure}[htb]
%	\centering
%	\includegraphics[height=6cm, keepaspectratio]{huber_Cluster 0.png}
%	\caption{Training History for Cluster 0 Model}
%	\label{fig:cluster_0_loss}
%\end{figure}
%%
%\begin{figure*}[htb]
%	\centering
%	\includegraphics[height=6cm, keepaspectratio]{huber_Cluster 1.png}
%	\caption{Training History for Cluster 1 Model}
%	\label{fig:cluster_1_loss}
%\end{figure*}
%%
%\begin{figure}[htb]
%	\centering
%	\includegraphics[height=6cm, keepaspectratio]{huber_Cluster 2.png}
%	\caption{Training History for Cluster 2 Model}
%	\label{fig:cluster_2_loss}
%\end{figure}
%
%\begin{figure*}[htb]
%	\centering
%	\includegraphics[height=6cm, keepaspectratio]{huber_Without Clustering.png}
%	\caption{Training History for Model Trained Without Clustering}
%	\label{fig:without_clustering_loss}
%\end{figure*}
%

\section{System Model}

Our proposed energy demand predictor system is shown in Fig.~\ref{fig:system_model}. We modeled the clients as smart homes capable of monitoring the energy usage across the household. The central server can be an electrical power company. The smart home system in each client is capable of training a neural network on its local data, but each individual client may have low amount of training data. Due to communication resource constraints, client households may also be unable to centralize their data or share among each other. To tackle these challenges, we trained our RNN model with the federated learning paradigm which allows clients to collaboratively train the model on their own local data without needing to share their data with a central entity. 

At the start of each federated training round, the central server or the power company clusters the participating client households based on the similarity of their housing attributes. After that, the server distributes a base global RNN model to all client households in each cluster. Each client individually trains the base RNN model on its own data.  After training the model, each client sends its model updates back to the server. The central server aggregates the model updates from the clients of the same cluster together to create a global model for each cluster. Then, the cluster specific global models are distributed back to the clients in the respective clusters. This process is independently repeated for each cluster until the respective global model converges.

A client can utilize the trained global model in its smart home system to predict the upcoming energy usage and effectively manage the energy consumption. When the electrical power company requires clients’ future energy demand, it can simply request the respective clients to make highly granular predictions on their local usage data without any need for data transmission.

%
%
%
%
%\begin{figure}[htbp]
%	\centering
%	\includegraphics[width=0.45\textwidth, keepaspectratio]{model_architecture_2.png}
%%	\includegraphics[height=7cm, keepaspectratio]{model_architecture_2.png}
%	\caption{Model Architecture}
%	\label{fig:nn_architecture}
%\end{figure}
%
%

\section{Federated Model Training and Client Clustering}

In a typical federated learning workflow \cite{Konecn2016FederatedLS,Konecn2016FederatedOD}, the server distributes a base global model~$\theta_t$ to the participating clients at the start of each training round~$t$. Each client $i$ trains the model on its local data~$D_i$ using stochastic gradient descent to generate model updates.
\begin{equation}
\Delta \theta_i^{t+1} = \textrm{SGD} (\theta^t, D_i) - \theta^t, \; i=1,...,m
\label{eqn:model_updates}
\end{equation}
Each client sends its model updates back to the server and the server aggregates the model updates by weighted averaging to produce the next global model.
\begin{equation}
\theta^{t+1} = \theta^t + \sum_{i=1}^m \frac{|D_i|}{|D|} \Delta \theta_i^{t+1}
\label{eqn:model_aggregation}
\end{equation}
Then, the global model is distributed back to the clients. Although the central server does not directly take part in the training process, it is responsible for monitoring and orchestrating the federated training and model aggregation process.

To increase the global model convergence rate, we used \mbox{OPTICS} (Ordering points to identify the clustering structure)~\cite{inproceedingsoptics} algorithm to cluster the clients with similar attributes. For each cluster, a global model is produced by aggregating the model updates from the cluster members. OPTICS is an unsupervised clustering algorithm with $\epsilon$ and $MinPts$ parameters. $\epsilon$ is the distance (radius) around a point to consider for clustering. $MinPts$ defines the minimum number of required points to form a cluster. If there are at least $MinPts$ points in $\epsilon$ radius neighborhood $N_\epsilon$ of a point $p$, then $p$ is a core point. 

%In OPTICS, the core distance which is the distance to the $MinPts^\textrm{th}$ closest point is defined by \eqref{eqn:core_distance}.
%%
%\begin{equation}
%%	\textrm{core-dist}_{\epsilon,MinPts} =
%\textrm{d}^\textrm{core}_{(p)} =
%\begin{cases}
%\textrm{UNDEFINED}, & \mbox{if } |N_\epsilon(p)|<MinPts  \\ \\
%MinPts\textrm{-th smallest} \\ \textrm{distance in } N_\epsilon(p), & \mbox{otherwise}
%\end{cases}
%\label{eqn:core_distance}
%\end{equation}
%%
%The reachability-distance of point $o$ from point $p$ is defined by \eqref{eqn:reachability_distance} which is the maximum between the core distance of $p$, and the distance between $p$ and $o$.
%%
%\begin{equation}
%%	\textrm{core-dist}_{\epsilon,MinPts} =
%\textrm{d}^\textrm{reach}_{(o,p)} =
%\begin{cases}
%\textrm{UNDEFINED}, & \mbox{if } |N_\epsilon(p)|<MinPts  \\ \\
%\textrm{max}(\textrm{d}^\textrm{core}_{(p)}, \textrm{d}(p, o)), & \mbox{otherwise}
%\end{cases}
%\label{eqn:reachability_distance}
%\end{equation}
Algorithm~\ref{algo:1} shows the details of out proposed method.
\begin{algorithm}
	\caption{Federated training with clustered aggregation}
	\label{algo:1}
	\floatname{algorithm}{Procedure}
	\begin{algorithmic}[1]
		\STATE \textbf{Input:} Attribute data from each of $m$ clients.
		\STATE Based on attributes, server groups clients into $n$ clusters.
		\STATE Server randomly initializes a base model $w_{rand}$.
		
		\FOR{each cluster $C^k$ with $k=1,2,...,N$, \textbf{in parallel}}
		\STATE Initialize cluster specific model weights, $w_k \gets w_{rand}$
		\ENDFOR
		
		\FOR{each cluster $C^k$ with $k=1,2,...,N$, \textbf{in parallel}}
		\FOR{communication round $t=1,2,....,T$}
		\FOR{each client $c^i$ in cluster $C^k$, $i=1,2,...,L$ \textbf{in parallel}}
		\STATE Synchronize local model with latest cluster specific model: $w_k^i \gets w_k$
		\STATE Update local model $w_k^i$ by training on local data, $D^i=\left\{ X^i,Y^i \right\}$
		\STATE Transmit updated $w_k^i$ back to the server.
		\ENDFOR
		\STATE Update model weights for cluster specific model by aggregation: $w_k \gets \frac{1}{l} \sum_{i=1}^{l} w_k^i$
		\ENDFOR
		\ENDFOR
		\STATE \textbf{Output:} cluster specific models with trained weights: $w_k$, $k=1,2,...,N$
	\end{algorithmic}
\end{algorithm}
%\dfrac{num}{den}
%
\begin{figure}[htbp]
	\centering
	\includegraphics[width=0.4\textwidth, keepaspectratio]{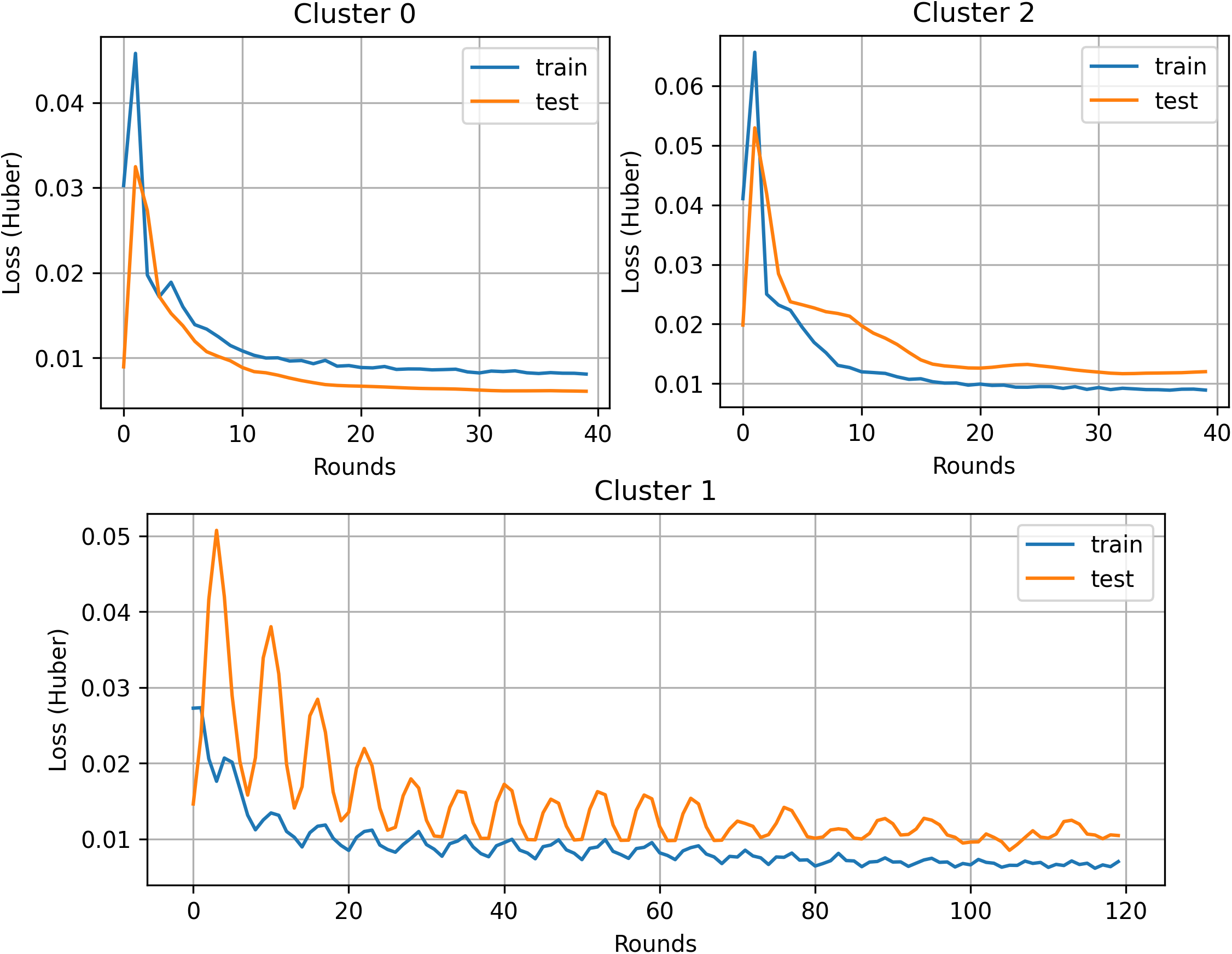}
	\caption{Training History for Cluster 0, 1 and 2 Models}
	\label{fig:training_with_clustering}
\end{figure}
\begin{figure}[htbp]
	\centering
	\includegraphics[width=0.4\textwidth,, keepaspectratio]{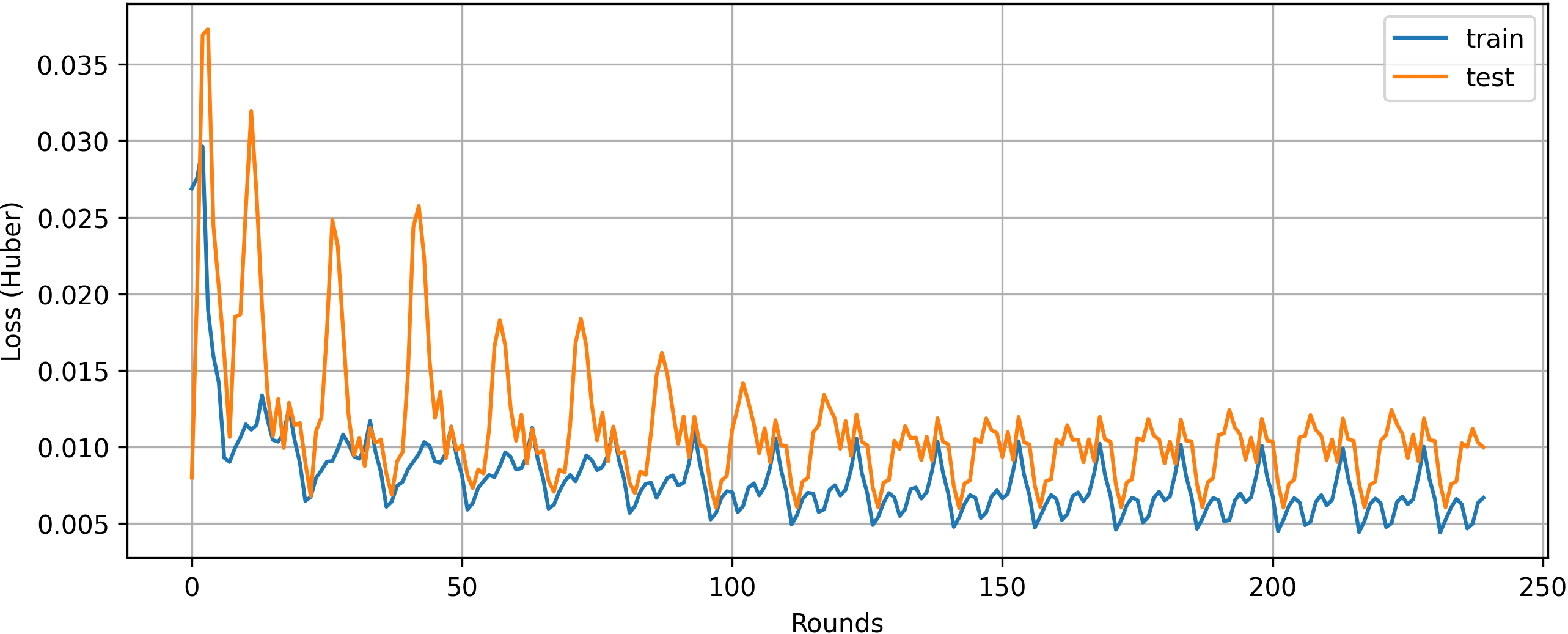}
	\caption{Training History for Model Trained Without Clustering}
	\label{fig:without_clustering_loss}
\end{figure}
\begin{figure}[htbp]
	\centering
	\subfloat[Cluster 0]{
		\label{subfig:tvp_cluster_0}
		\includegraphics[width=0.45\textwidth, keepaspectratio]{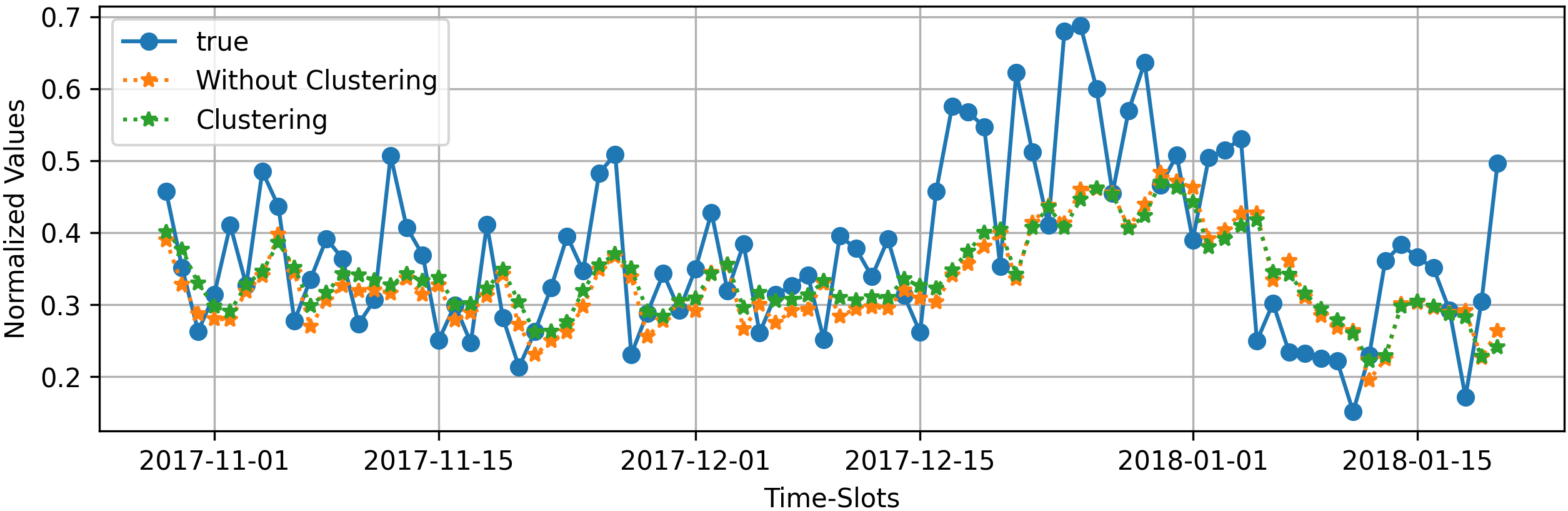} } 
	
	\subfloat[Cluster 1]{
		\label{subfig:tvp_cluster_1}
		\includegraphics[width=0.45\textwidth, keepaspectratio]{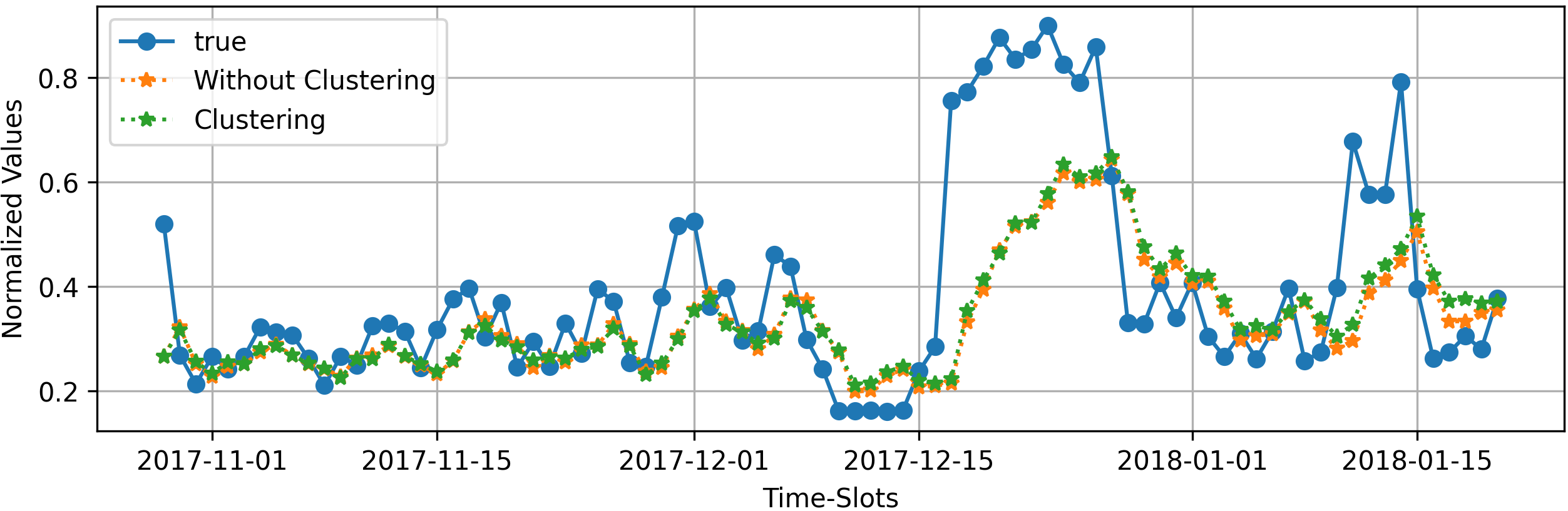} } 
	
	\subfloat[Cluster 2]{
		\label{subfig:tvp_cluster_2}
		\includegraphics[width=0.45\textwidth, keepaspectratio]{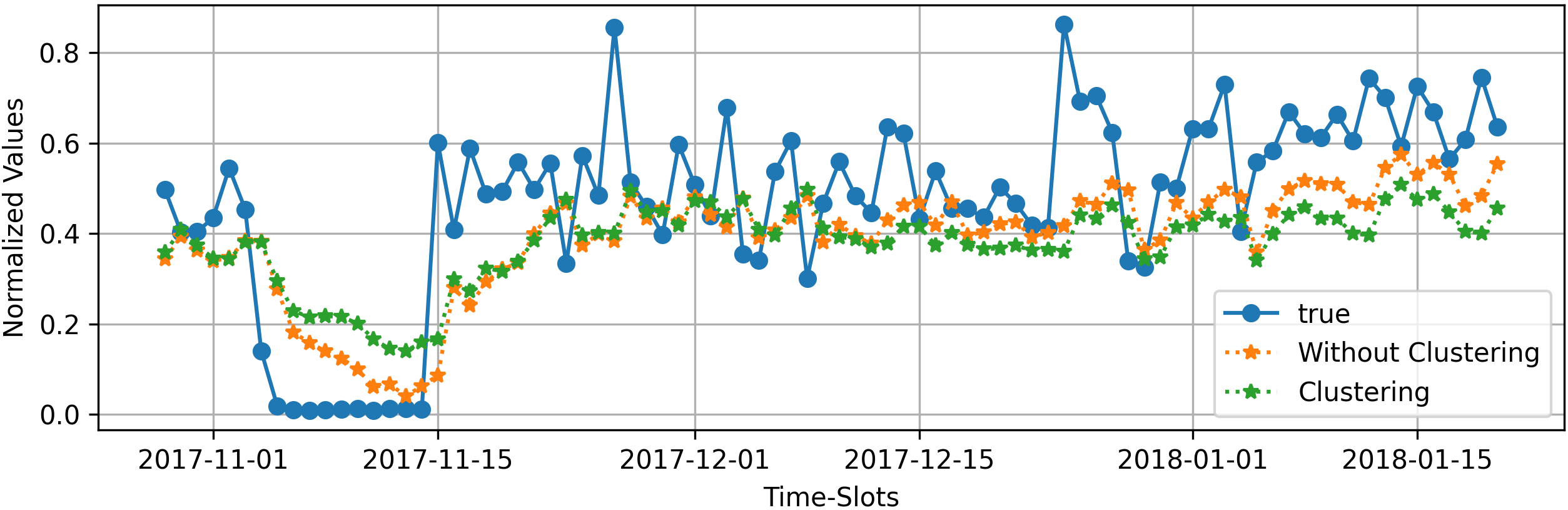}} 
	
	\caption{Predictions from Cluster Specific Models and Model Trained Without Clustering}
	\label{fig:tvp}
	
\end{figure}

%\begin{figure}[htbp]
%	\centering
%	\includegraphics[width=0.48\textwidth, keepaspectratio]{train_and_test_histogram_cluster_0.png}
%	\caption{Training and Testing Data Distributions (Cluster 0)}
%	\label{fig:data_distributions_cluster_0}
%\end{figure}

%
%

\section{Experimental Settings}

For our experiments, we used “HUE: The Hourly Usage of Energy Dataset for Buildings in British Columbia” \cite{DVN/N3HGRN_2018} dataset. It contains almost three years worth of hourly energy usage data for twenty-two households as well as the housing attributes. We limited the time period for training data between `2015-09-29' and `2017-09-29' and based on the availability, houses with IDs \mbox{3-15, 19, and 20} are chosen. Available energy readings after `2017-09-29' for each house are kept as hold out test data. Hourly energy usage readings are re-sampled into daily energy usage data.

Initially, we input housing attributes into the OPTICS algorithm and to get the clustered house IDs as output. Types of attributes used for clustering include: house type (eg.bungalow or apartment), facing direction, region, rental units, heating type (electricity or gas). In our experiments, we used scikit-learn \cite{scikit-learn} implementation of OPTICS algorithm and we set the minimum number of cluster samples as two and other parameters are left as default. On our clients, clustering produced three clusters and a set of clients are sampled as noise. Cluster 1 includes house IDs 3, 5, and 14. Cluster 2 includes house IDs 4, 8, 9, 10, 13 and 19. Lastly, cluster 3 includes house IDs of 6 and 12. For each of the three clusters, we train an RNN model in a federated learning manner.
%
%\begin{table}
%	\caption{Housing Attributes}
%	\begin{center}
%		\bgroup
%		\def\arraystretch{1.1}%
%		\begin{tabularx}{0.8\columnwidth}{|p{2cm}|X|}
%			\hline
%			\textbf{Property} & \textbf{Description}\\
%			\hline
%			House Type & Type of the household such as a bungalow or an apartment. \\
%			\hline
%			Facing Direction & The direction the house is facing. \\
%			\hline
%			Region & Region where the house is located. \\
%			\hline
%			Rental Units & Number of rental suites in the house. \\
%			\hline
%			Heating Type & Electricity or gas, which is used as the main source of heating.\\
%			\hline
%		\end{tabularx}
%		\egroup
%		\label{table:household_meta_information}
%	\end{center}
%\end{table}
%
%\begin{table}
%	\caption{Clustered Client Households}
%	\begin{center}
%		\bgroup
%		\def\arraystretch{1.1}%
%		\begin{tabularx}{0.8\columnwidth}{|p{2cm}|X|}
%			\hline
%			\textbf{Cluster ID} & \textbf{House IDs}\\
%			\hline
%			0 & 3, 5, 14 \\
%			\hline
%			1 & 4, 8, 9, 10, 13, 19 \\
%			\hline
%			2 & 6, 12 \\
%			\hline
%			Noise & 7, 11, 15, 20 \\
%			\hline
%		\end{tabularx}
%		\egroup
%		\label{table:clustered_households}
%	\end{center}
%\end{table}
%

We use Keras API \cite{chollet2015keras} with TensorFlow \cite{tensorflow2015-whitepaper} backend for implementing the RNN model. Bidirectional LSTM (long short-term memory)~\cite{10.1162/neco.1997.9.8.1735} layers are used for constructing the RNN model. Since LSTM  layers can recognize patterns of long-term dependencies, it is suitable for predicting time series data. In a bidirectional LSTM layer \cite{650093}, neurons are split into two parts in which one part is responsible for the forward states and the other for the backward states. A bidirectional LSTM layer can utilize both past and future information of an input time series for training. Our RNN model is a sequential model that contains two bidirectional LSTM layers, two dense layers and an output dense layer. Our RNN model is configured to receive thirty days’ worth of energy usage data and predict the next day demand. 

RNN models for each cluster are trained separately in federated learning paradigm. At the start of every training round, the latest global model is synchronized to the participating clients as the base model. In every training round each client trains the base model for five epochs. At the end of every training round, the resulting global model is evaluated on the test data of its respective cluster.

\section{Evaluation and Results}

\begin{table}
	\caption{Test Loss Comparison}
	\begin{center}
		\bgroup
		\def\arraystretch{1.1}%
		\begin{tabular}{*6{|c}|} \cline{3-6}
			
			\multicolumn{2}{c|}{} & \multicolumn{2}{c|}{\thead{Cluster Specific \\ Model}} &  \multicolumn{2}{c|}{\thead{Model Trained \\ Without Clustering}}\\ \hline
			
			\thead{Cluster \\ ID} & \thead{House \\ ID} & Rounds & \thead{Test Loss \\ (Huber)} & Rounds & \thead{Test Loss \\ (Huber)}\\ \hline
			
			\multirow{3}{*}{0} & 3 & \multirow{3}{*}{40} & 0.00818 & \multirow{11}{*}{240} & 0.00861 \\
			& 5	& & 0.00506 & & 0.00516 \\
			& 14 & & 0.00519 & & 0.00546 \\ \cline{1-4} \cline{6-6}
			
			\multirow{6}{*}{1} & 4 & \multirow{6}{*}{120} & 0.01060 & & 0.01245 \\
			& 8	& & 0.01398 & &0.01333 \\
			& 9	& & 0.00953 & & 0.00955 \\
			& 10 & & 0.00844 & & 0.00840 \\
			& 13 & & 0.01295 & & 0.01315 \\
			& 19 & & 0.00766 & & 0.01241 \\ \cline{1-4} \cline{6-6}
			
			\multirow{2}{*}{2} & 6 & \multirow{2}{*}{40} & 0.01817 & & 0.01336 \\
			& 12 & & 0.00702 & & 0.00452 \\
			\hline
		\end{tabular}
		\egroup
		\label{table:test_loss}
	\end{center}
\end{table}
%
%\begin{figure*}
%	\begin{subfigure}{0.31\textwidth}
%		\includegraphics[width=\linewidth]{huber_Cluster 0.png}
%		\caption{First subfigure} \label{fig:1a}
%	\end{subfigure}%
%	\hspace*{\fill}   % maximize separation between the subfigures
%	\begin{subfigure}{0.31\textwidth}
%		\includegraphics[width=\linewidth]{huber_Cluster 0.png}
%		\caption{Second subfigure} \label{fig:1b}
%	\end{subfigure}%
%	\hspace*{\fill}   % maximizeseparation between the subfigures
%	\begin{subfigure}{0.31\textwidth}
%		\includegraphics[width=\linewidth]{huber_Cluster 0.png}
%		\caption{Third subfigure} \label{fig:1c}
%	\end{subfigure}
%	
%	\caption{A figure that contains three subfigures} \label{fig:1}
%\end{figure*}
%%

Federated training history plots for each cluster are shown in Fig.~\ref{fig:training_with_clustering}. For comparison, we also trained an RNN model in the federated learning manner on all available clients without clustering and the history plot is shown in Fig.~\ref{fig:without_clustering_loss}. Cluster specific models for cluster 0 and 2 converged after training for forty rounds and that for cluster 1 converged after training for a hundred and twenty rounds. The model with all available clients is trained for two hundred and forty training rounds. Table~\ref{table:test_loss} shows the evaluation results the cluster specific models and the model trained without clustering, on corresponding clients' test data. Cluster specific models are evaluated on the test data of clients from the respective clusters while the model trained on all available clients is evaluated on all clients' test data. In Table~\ref{table:test_loss}, we can observe that there is no significant performance loss between cluster specific models which are trained with a relatively fewer number of participating clients and the model trained without clustering where all clients participate. From the Fig.~\ref{fig:training_with_clustering} and \ref{fig:without_clustering_loss}, it is evident that cluster specific models achieved the optimum significantly faster compared to the single federated model trained with all clients. In Fig.~\ref{fig:without_clustering_loss}, we can observe that training and testing losses for the model trained without clustering slowly felled, but they never fully flattened out because different client households have different housing attributes and thus diverse energy usage patterns. In Table~\ref{table:test_loss}, for cluster 2, we can observe that the model trained without clustering performs slightly better than the cluster 2 specific model. This is because the cluster 2 specific model is trained on only two clients and the amount of training data was significantly less than that is available for the model trained with all clients. Fig.~\ref{fig:tvp} visualizes the predictions between the cluster specific models and the model trained without clustering on test samples of a client from each of the cluster. 

%In the training history for cluster 0 of Fig.~\ref{fig:training_with_clustering}, we observe that testing results are slightly better than the training results when training the model for cluster 0. This is because distribution of training data for cluster 0 is slightly different from the testing data which is shown in Fig.~\ref{fig:data_distributions_cluster_0}.
%

%\begin{figure}[H]
%	\centering
%	\subfloat[Training Data Distribution]{
%		\label{subfig:Training Data Distribution}
%		\includegraphics[]{train_and_test_histogram_cluster_0_a.png} } 
%	
%	\subfloat[Testing Data Distribution]{
%		\label{subfig:Testing Data Distribution}
%		\includegraphics[]{train_and_test_histogram_cluster_0_b.png} } 
%
%	\caption{Training and Testing Data Distributions (Cluster 0)}
%	\label{fig:data_distributions_cluster_0}
%	
%\end{figure}

%
\section{Conclusion}

In this paper, we studied the process of developing an RNN based energy demand predictor system by training it using clustering and federated learning based methods. Federated learning allows us to take advantage of distributed data in clients by training local models without the need for data transmission. But different clients can have different data distributions and federated training process of a model can take a long time to converge. To speed up the convergence rate of the global model, clients with similar attributes can be clustered together and the model updates from the clients' of the same clusters can be aggregated together. We experimented by training an RNN model for each cluster of clients and compare the results with a single RNN model trained for all available clients. Our method can reduce the communication cost required for collecting training data onto a server since the local data never left the clients. And from our experiments, we observe that by clustering clients, cluster specific models converge significantly faster compare to a model trained without clustering.

%
%\begin{figure*}
%	\begin{subfigure}{0.31\textwidth}
%		\includegraphics[width=\linewidth]{huber_Cluster 0.png}
%		\caption{First subfigure} \label{fig:1a}
%	\end{subfigure}%
%	\hspace*{\fill}   % maximize separation between the subfigures
%	\begin{subfigure}{0.31\textwidth}
%		\includegraphics[width=\linewidth]{huber_Cluster 0.png}
%		\caption{Second subfigure} \label{fig:1b}
%	\end{subfigure}%
%	\hspace*{\fill}   % maximizeseparation between the subfigures
%	\begin{subfigure}{0.31\textwidth}
%		\includegraphics[width=\linewidth]{huber_Cluster 0.png}
%		\caption{Third subfigure} \label{fig:1c}
%	\end{subfigure}
%	
%	\caption{A figure that contains three subfigures} \label{fig:1}
%\end{figure*}

\nocite{*}
\footnotesize
\bibliographystyle{IEEEtran}
\bibliography{reference}

\end{document}